\documentclass{article} 
\usepackage{iclr2021_conference,times}

\usepackage{amsmath,amsfonts,bm}

\def\eqref#1{equation~\ref{#1}}

\def\1{\bm{1}}

\DeclareMathAlphabet{\mathsfit}{\encodingdefault}{\sfdefault}{m}{sl}
\SetMathAlphabet{\mathsfit}{bold}{\encodingdefault}{\sfdefault}{bx}{n}

\usepackage{graphicx}
\usepackage{hyperref}
\usepackage{url}
\usepackage{amssymb}

\title{Privacy-aware Early Detection of COVID-19 through Adversarial Training}

\author{%
    Omid Rohanian\thanks{Correspondence to: \texttt{omid.rohanian@eng.ox.ac.uk}, \texttt{samaneh.kouchaki@surrey.ac.uk}, \texttt{Andrew.Soltan@cardiov.ox.ac.uk}} \textsuperscript{\hspace{0.3em}$1$}
    \And
    Samaneh Kouchaki\textsuperscript{$1, 2, 3$}
    \And
    Andrew Soltan\textsuperscript{$4, 5$}
    \And
    Jenny Yang\textsuperscript{$1$}
    \And
    Morteza Rohanian\textsuperscript{$6$}
    \And
    Yang Yang\textsuperscript{$1, 7$}
    \And    
    David Clifton\textsuperscript{$1$, $7$}
    \and\\
    \centerline{$^1$ Department of Engineering Science, University of Oxford, Oxford, UK}\\
    \centerline{$^2$ CVSSP, University of Surrey, Surrey, UK}\\
    \centerline{$^3$UK Dementia Research Institute Care Research and Technology Centre,} \\ {Imperial College London and the University of Surrey, UK}\\
    \centerline{$^6$ Queen Mary University of London, London, UK}\\
    \centerline{$^7$ Oxford-Suzhou Centre for Advanced Research, Suzhou, China}\\
    \centerline{$^5$ John Radcliffe Hospital, Oxford University Hospitals NHS Foundation Trust, UK}\\
    \centerline{$4$ RDM Division of Cardiovascular Medicine, University of Oxford, UK}\\

\vspace{-16pt}
}

\iclrfinalcopy 
\begin{document}

\maketitle

\begin{abstract}
Early detection of COVID-19 is an ongoing area of research that can help with triage, monitoring and general health assessment of potential patients and may reduce operational strain on hospitals that cope with the coronavirus pandemic. Different machine learning techniques have been used in the literature to detect potential cases of coronavirus using routine clinical data (blood tests, and vital signs measurements). Data breaches and information leakage when using these models can bring reputational damage and cause legal issues for hospitals. In spite of this, protecting healthcare models against leakage of potentially sensitive information is an understudied research area. In this work, we examine two machine learning approaches, intended to predict a patient's COVID-19 status using routinely collected and readily available clinical data. We employ adversarial training to explore robust deep learning architectures that protect attributes related to demographic information about the patients. The two models we examine in this work are intended to preserve sensitive information against adversarial attacks and information leakage. In a series of experiments using datasets from the Oxford University Hospitals (OUH), Bedfordshire Hospitals NHS Foundation Trust (BH), University Hospitals Birmingham NHS Foundation Trust (UHB), and Portsmouth Hospitals University NHS Trust (PUH) we train and test two neural networks that predict PCR test results using information from basic laboratory blood tests, and vital signs performed on a patients' arrival to hospital. We assess the level of privacy each one of the models can provide and show the efficacy and robustness of our proposed architectures against a comparable baseline. One of our main contributions is that we specifically target the development of effective COVID-19 detection models with built-in mechanisms in order to selectively protect sensitive attributes against adversarial attacks. 

\end{abstract}

\section{Introduction}
COVID-19 has impacted millions across the world. Its early signs cannot be easily distinguished from other respiratory illnesses and hence an accurate and rapid testing approach is vital for its management. RT-PCR assay of nasopharyngeal swabs is a widely accepted gold-standard test, which has several limitations, including limited sensitivity and slow turnaround time (12-24h in hospitals in high and middle-income countries). Several other techniques, including qualitative rapid-antigen tests (`lateral flow'; LFTs), point-of-care PCR, and loop mediated isothermal amplification have been proposed and are in various stages of validation and implementation~\citep{assennato2020performance,DHSC}. Among these techniques, lateral flow tests are favoured as they are inexpensive and do not require specialised laboratory equipment which allow for decentralised testing and faster results. However, sensitivity results for lateral flow testing vary greatly amongst groups, with reported values ranging from $40\%$ to $70\%$.~\cite{dinnes2021rapid,DHSC}. There are also numerous studies based on radiological imaging, including CT~\cite{khuzani2021covid}. Such tests are less widely available, involve a longer turnaround time, and expose patients to ionising radiation. 

There are a number of research studies on the deployment of machine learning techniques to detect COVID-19 from various widely available features, including demographic and laboratory markers~\citep{goodman2020machine, zoabi2021machine}. Inclusion of demographics in learning might lead to the development of biased tests, and even when they are not explicitly included in the feature representation, these attributes can potentially confound the model through their correlation with other features. We recently introduced a machine learning test based on vital signs, routine laboratory blood tests and blood gas~\citep{soltan2021rapid}. A strength of our test is the use of clinical data which is typically available within 1h, much sooner than the typical turnaround time of RT-PCR testing (up to 24h in hospitals in high- and middle- income countries). Current tests that employ machine learning are promising as they alleviate the need for specialised equipment, can potentially be more sensitive, and are faster than existing tests. Nonetheless they suffer from several shortcomings:
\begin{enumerate}
\item Most approaches that have appeared in the literature so far are based on basic machine learning techniques that require a complete retraining anytime a new batch of data is available. However, in a dynamic situation like a pandemic where new streams of data need to be processed, it is vital to incrementally learn from data without the need to start over and retrain the system using all the seen instances.
\item ML-based models explored in the COVID-19 literature are not equipped with an inherent mechanism to guard against possible issues that might arise due to the presence of demographic features. For example, models could easily get biased to a certain demographic group causing incorrect associations and overfitting.
\item Another issue is preserving the privacy of the patients and robustness against adversarial attacks. Most basic models can easily `leak' information, making it easy for an adversary to recover sensitive information contained in the hidden representation. As blood tests are known to include features which typically correlate with demographic features, such as sex and ethnicity, exclusion of demographics does not necessarily solve the problem. For example, health issues like Benign Ethnic Neutropenia \citep{haddy1999benign} or Sickle Cell Disease \citep{REES20102018} are predominantly found in a certain number of ethnic groups and much less likely to occur in others. As an additional example, healthy men and women have different reference ranges for blood tests \citep{park2016establishment}.
\end{enumerate}

This work aims to address the above-mentioned shortcomings in existing research. The proposed adversarial architectures (Section \ref{sec:methodology}) are designed to prevent the learning model from potentially encoding unwanted demographic biases and protect its sensitive information during the learning process. In the first architecture (Section \ref{ADV-Demographic}), protection of attributes is explicit, with the option to select the attributes for guarding against adversarial attacks. We will investigate in Section \ref{sec:demographic-crosstest} whether these direct protective measures would hurt generalisibility to unseen data. In the second architecture (Section \ref{adversarial_perturbation}), protecting attributes is based on a general adversarial regularisation and is not tied to any specific subset of selected attributes. 

Several recent studies in the field of natural language processing (NLP) have shown that textual data carries informative features regarding authors' race, age and other social factors. This makes embedding and predictive models susceptible to a wide range of biases that can negatively affect performance and severely limit generalisability. This kind of bias also raises concerns in areas where fairness and privacy are important. Numerous works have focused on the different ways representation learning can be biased to or against certain demographics and different countermeasures have been proposed to counteract bias \citep{gonen2019lipstick}. 
Most of these studies, however, are done using text and image data. 
Currently, there is limited research on the application of representation learning and adversarial models for healthcare applications.  

The proposed models in this study are designed to preserve sensitive information against adversarial attacks, allow incremental learning, and reduce the potential impact of demographic bias. However, the main focus of the work is in privacy preservation. The contributions of this work are as follows:

\begin{itemize}
  \item We introduce two adversarial learning models for the task of COVID-19 identification based on Electronic health records (EHR) that perform satisfactorily on a real COVID-19 dataset and in comparison with strong baselines. Unlike conventional tree-based methods, these architectures are well-suited for transfer learning, multi-modal data, and other advantages of neural models without a significant performance trade-off.     
  \item The models use adversarial regularisation to make them robust against leakage of sensitive information and adversarial attacks, which makes them suitable for scenarios where preservation of privacy is important or classification bias is costly. 
  \item We run a series of tests to quantitatively demonstrate the efficacy of the proposed architectures in protecting sensitive information against adversarial attacks in comparison with a neural model that is not adversarially trained.
  \item We perform several tests to observe the effect of this type of training on generalisability across different demographic groups.
  \item We externally validate the models using data from other hospital groups. 
\end{itemize}

\section{Privacy Attacks in Machine Learning and Healthcare}
\label{sec:priv-attacks}

\begin{figure}[ht]
\centering
\includegraphics[width=8.0cm]{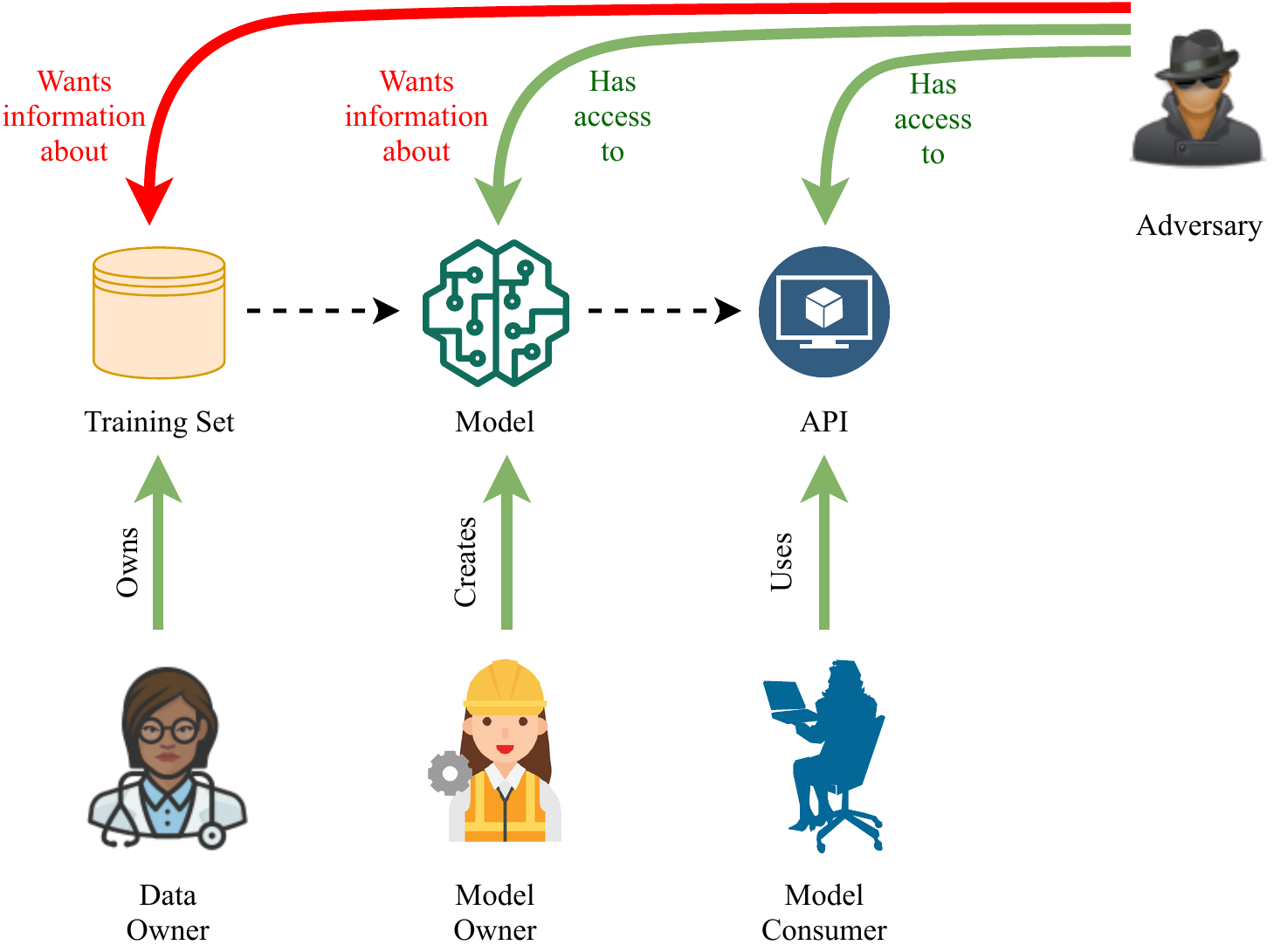}
\caption{Schematic view of privacy attacks for a machine learning model. Dashed lines represent information flow, and full lines signify possible actions.}
\label{attack-model}
\end{figure}

There are various ways a trained model can be attacked by an adversary. The goal in most of them is to infer some kind of knowledge that is not originally meant to be shared or is unintentionally encoded by the model. At least three different forms of attack are known, namely, membership inference, property inference, and model inversion \citep{shokri2017membership}. In this work, we focus on property inference, in which an adversary who has access to model's parameters during training, tries to extract information about certain properties of the training data that are not necessarily related to the main task. Figure \ref{attack-model} shows the general overview of privacy attacks according to \citet{rigaki2020survey}. The adversary, in our case, can see the model and its parameters and wants information about the data to which they do not have direct access to.  

Attacks of this kind are possible in any scenario where the model is stored and trained on an external server. Protecting an ML model against property inference attacks is especially useful in the context of collaborative and federated learning, where models locally train on different portions of the dataset and share their parameters over a network that might or might not be fully secure against eavesdropping \citep{melis2019exploiting}. 

Within the context of healthcare, such attacks can reveal sensitive personal data and prove disastrous for hospitals. GDPR defines personal data as `any information relating to an identified or identifiable natural person'. Article 9(1) of the GDPR declares the following types of personal data as sensitive: data revealing racial or ethnic origin, political opinions, religious or philosophical beliefs, or trade union membership, genetic and biometric data, and data concerning health or sex life or sexual orientation of the subject \citep{voigt2017eu}. 

Sensitive information such as age, gender, location, or ethnicity are usually quantised or anonymised in large healthcare datasets. However, as we will see in Section \ref{attacking-models}, this information can be easily recovered by a simple attack model because of the implicit associations that exist between such information and other features in the dataset.  

Property inference attacks are not limited to recovering any specific type of data and can predict both categorical and numerical values. For instance, they can be used to train attacker models that learn to identify both demographic features (implicitly present in the data) and blood test features (explicitly present) that highly correlate with certain diseases. It is then possible to use this trained model to re-identify some patients based on their demographic features and possible combination of diseases \citep{jegorova2021survey}. 

\section{Task Definition}
\label{sec:task-def}

In our binary classification setting, each neural network $f$ is trained to predict labels $y_{1}$, $y_{2}$, ..., $y_{n}$ from instances $x_{1}$, $x_{2}$, ..., $x_{n}$. Each instance $x_{i}$ contains a set of sensitive (in this case demographic) discrete features $z_{i} \in {1,2,...,k}$ which we intend to ``protect" \footnote{We would ideally want the transformation $y_{i} = f(x_{i})$ not to be confounded by specific values of $z_{i}$. However, our experiments here are focused on privacy preservation and not on the closely related subject of debiasing.}. These sensitive features are called \textit{protected attributes}. 

In the context of classification, any neural network $f(x)$ can be characterised as an encoder, followed by a linear layer $W: f(x) = W \times h(x)$. $W$ can be seen as the last layer of the network (i.e. dense + softmax) and $h$ is all the preceding layers \citep{ravfogel2020null}.

Suppose we have an \textit{attacker} model $f_{att}$ that is trained on the encoder $h(x)$ of a neural classifier in order to predict $z_{i}$. If this trained adversary is able to predict $z_{i}$ based on the encoded representation from the model, the model has \textit{leaked} and privacy of the model has been compromised.  

It is unlikely that $h(x)$ would be completely guarded against an attack. If it encodes sufficient information about $x_{i}$ it might reveal some information to a properly trained $f_{att}$. We say that the trained model $f$ is private with regards to $z_{i}$ if an attacker model $f_{att}$ that has access to $f$'s  encoder ($h(x)$) cannot predict $z_{i}$ with a greater probability than a majority class baseline.  

If we perturb $h(x)$ too much, it will not be informative to $f_{att}$ but would also fail in accurately predicting the main task label $y_{i}$. Therefore, we would like to ensure privacy against potential attackers with regards to the protected attributes while achieving a reasonably good result in the main task.   

\section{Methodology}
\label{sec:methodology}

We follow a standard supervised learning scenario where each training instance $x_{i}$ represents information from blood tests and vital signs for each patient seen at the hospital and $y_{i}$ is the corresponding Boolean value denoting the result of the PCR test for that patient. The task is to train a model to predict the correct label for each patient. 

\subsection{Adversarial training based on gradient reversal}
\label{ADV-Demographic}
The first adversarial architecture we explore is comprised of one main part and a number of secondary networks: 

\renewcommand{\theenumi}{\Roman{enumi}}
\begin{enumerate}
    \item A main classifier $M$ that is the central component of the model. It consists of a stack of $n$ fully connected layers with dropout and batch normalisation, followed by a softmax layer at the end.

    \item $d$ networks with auxiliary objectives separate from the main task. Supposing we have $d$ categorical features, each of these secondary networks  (henceforth referred to as discriminators) predict the value for that feature given each training instance. 
\end{enumerate}

\begin{figure}[ht]
\centering
\includegraphics[width=6cm]{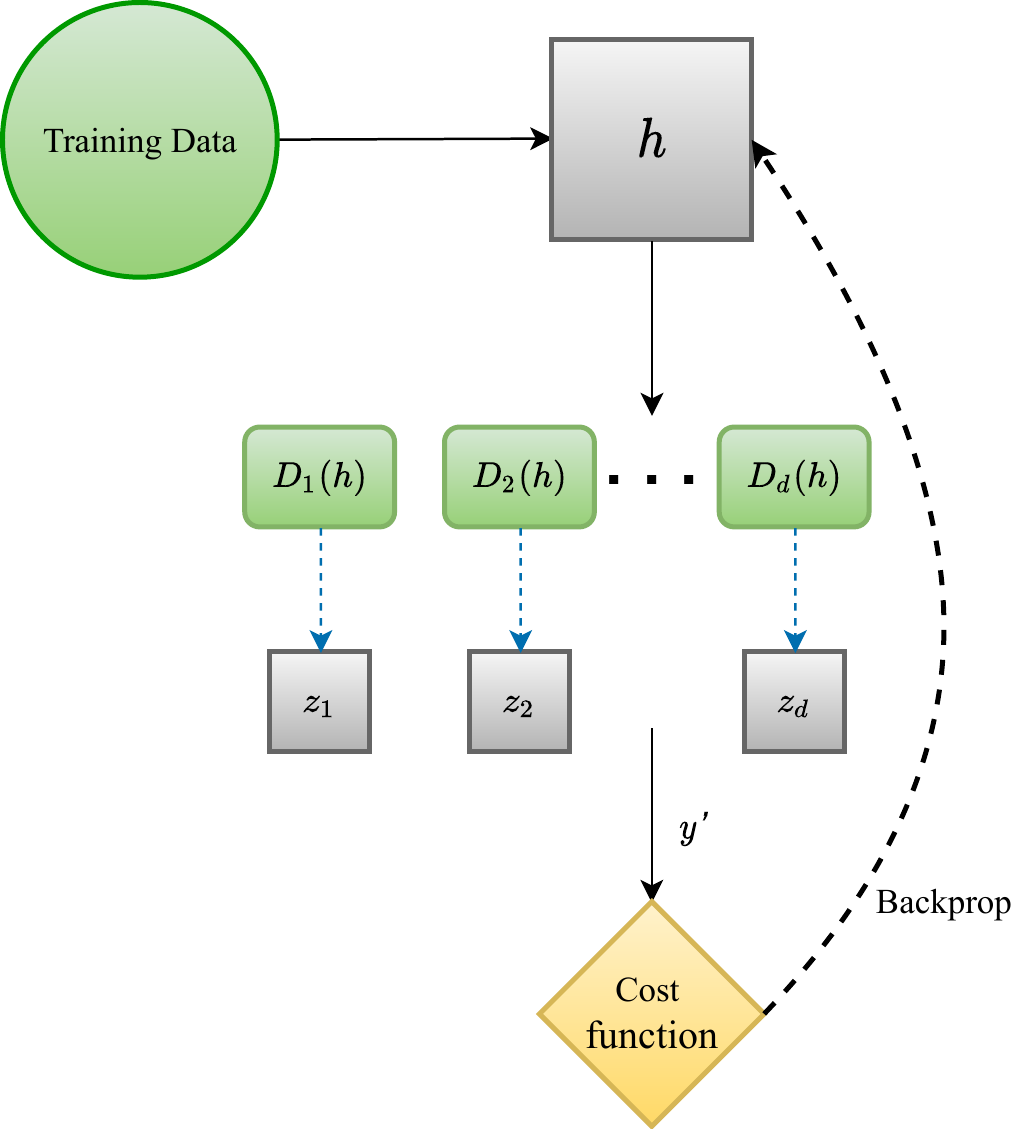}
\caption{Overall structure of the proposed model. Each $D_i$ is a discriminator that aims to predict any of the $d$ categorical features $z_i$}
\label{demogstructure}
\end{figure}

Assume $h_i$ is the representation of an instance at the \textit{i}th layer within $M$. This is the point of interception where the auxiliary networks get access to the contents of $M$. All these components then train in tandem with the following loss function:
\begin{equation}\label{loss_fn}
    L = L_M - \sum_{i=1}^{d}\lambda L_{D_{i}}
\end{equation}

Each $D_{i}$ corresponds to a separate discriminator network that predicts one of the $d$ different categorical features of interest. $\lambda$ is a weighting factor and can control the contribution of each individual auxiliary loss. Formula \ref{loss_fn} is set up so that after backpropagation, the contents of $h$ be maximally informative for the main task, and minimally informative for prediction of the protected features. Loss of the main task is computed using binary cross entropy. 

If $x$ and $y$ are the features and labels, $\hat{y}$ and $\hat{z}$ the predictions for the main target and protected features, $\theta_{M}$ and $\theta_{D_i}$ the parameters of the main classifier and its $d$ discriminators, and $L$ is the joint binary cross entropy loss function, we can formulate the training objective as finding the optimal parameters $\hat{\theta}$ such that: 

\begin{equation}\label{objective}
    \hat{\theta} = min_{\theta_{M}} max_{\left \{ \theta_{D_{i}} \right \}_{i=1}^{d}} L (\hat{y}(x;\theta_{M}),y) - \lambda \sum_{i=1}^{d} L (\hat{z}(x;\theta_{D_{i}}),z_{i})
\end{equation}

\subsubsection{Gradient Reversal Layer}
\label{GRL-layer}

As discussed in Section \ref{ADV-Demographic}, during training, the objective is to jointly minimise both of the following terms \footnote{Our formulation of GRL in this section is based on \citet{elazar2018adversarial}}:

\begin{equation} 
\label{eql1}
    arg \ \min_{D} L(D(h(x_{i})),z) 
\end{equation} 
\begin{equation}
\label{eql2}
arg \ \min_{h, c} L(c(h(x_{i})),y_{i}) - L(D(h(x_{i})),z)
\end{equation} 

where each $x_{i}$ is an instance of the data which is associated with the protected attribute $z$. $D$ is the discriminator (the adversarial network), and $c$ is the classifier used to predict the labels for the main task from representation $h$. $L$ denotes the loss function. 

Using an optimisation trick called the \textit{Gradient Reversal Layer (GRL)}, we can combine the above terms into a single objective. This idea was first introduced in the context of domain adaptation \citep{ganin2015unsupervised} and was later also applied to text processing \citep{elazar2018adversarial,li2018towards}. GRL is easy to implement and requires adding a new layer to the end of the Discriminator's encoder. 

During forward propagation, GRL acts as an identity layer, passing along the input from the previous layer without any changes. However, during backpropagation, it multiplies the computed gradients by $-1$. Mathematically this layer can be formulated as a pseudo\-function with the following two incompatible equations:

\begin{equation}\label{grl-math}
    \begin{cases}
      GRL(x) = x & \text{if in forward mode}\\
      \frac{dGRL(x)}{dx} = - I & \text{if in backprop mode}\\
    \end{cases}       
\end{equation}

Using this layer, we could formulate the loss function into one single formula, and perform a single backpropagation in each training epoch. For the trivial case of having only one protected attribute, we can consolidate equations \ref{eql1} and \ref{eql2} with the following: 

\begin{equation}\label{grl_onez}
    arg \ \min_{h,c,D} L(c(h(x_{i})),y_{i})+L(D(\lambda GRL(h(x_{i}))),z)
\end{equation}

The objective is to minimise the total loss, and for the case of the discriminator, the gradients are reversed and scaled by $\lambda$. We can generalise this to the case where we have multiple (in our case $3$; namely, age, gender, and ethnicity) protected attributes and corresponding $D_{i}$s:

\begin{equation}\label{grl_threez}
L = L_M + \sum_{i=1}^{d} L(D_{i}(\lambda GRL(h(x),z_{i})))))
\end{equation}

\subsection{Adversarial training based on fast gradient sign method}
\label{adversarial_perturbation}

As the second adversarial architecture, we develop another model in which the adversarial component can perturb the representation during training with some added noise. The direction of this noise (i.e. whether the added noise is a positive or negative number) is dependent on the signs of the computed gradients.

This adversarial method is based on linear perturbation of inputs fed to a classifier. In every dataset, the measurements enjoy a certain degree of precision, below which could be considered negligible error $\epsilon$. If $x$ is the representation of an instance, it is likely that the classifier would treat $x$ the same as $\tilde{x}=x+\eta$, as long as $\left \| \eta \right \|_{\infty} < \epsilon$.   

However, this small perturbation grows when it is multiplied by a weight matrix $w$:

\begin{equation}\label{perturb-weight}
   w^{\top}\tilde{x} = w^{\top}(x+\eta) = w^{\top}x+w^{\top}\eta
\end{equation}

The perturbation is maximised when we set $\eta = sign(w)$, predicated on the assumption that it remains within the max-norm constraint defined above. In the context of deep learning, the method can be formulated in the following way:

If $\theta$ is the parameters of the model, and $J$ is the cost function, during training, for each instance a perturbation of $\eta$ is added to the representation of the instance such that:

\begin{equation}\label{eta}
   \eta = \epsilon sign(\triangledown_{x} J(\theta,x,y_{pred}))
\end{equation}

This procedure is known as the \textit{fast gradient sign method (FGSM)}, originally introduced in a seminal $2015$ paper by \citet{goodfellow2015explaining}. It can be viewed either as a regularisation technique or a data augmentation method that includes unlikely instances in the dataset. For training, the following adversarial objective function can be used:

\begin{equation}\label{fgsm-cost}
   \tilde{J}(\theta,x,y_{pred}) = \alpha J(\theta,x,y)+(1-\alpha)J(\theta,x + \epsilon sign(\triangledown_{x} J(\theta,x,y_{pred})))
\end{equation}

This method can be seen in terms of making the model robust against worst case errors when the data is perturbed by an adversary \citep{goodfellow2015explaining}. Because of this regularisation, our expectation is that hidden representations would become less informative to an attacker network that attempts to retrieve demographic attributes. Following the original paper, $\alpha$ is usually taken to be $0.5$, which turns the equation into a linear combination with equal weights given to both terms in the objective function. 

\begin{figure}[ht]
\centering
\includegraphics[width=6cm]{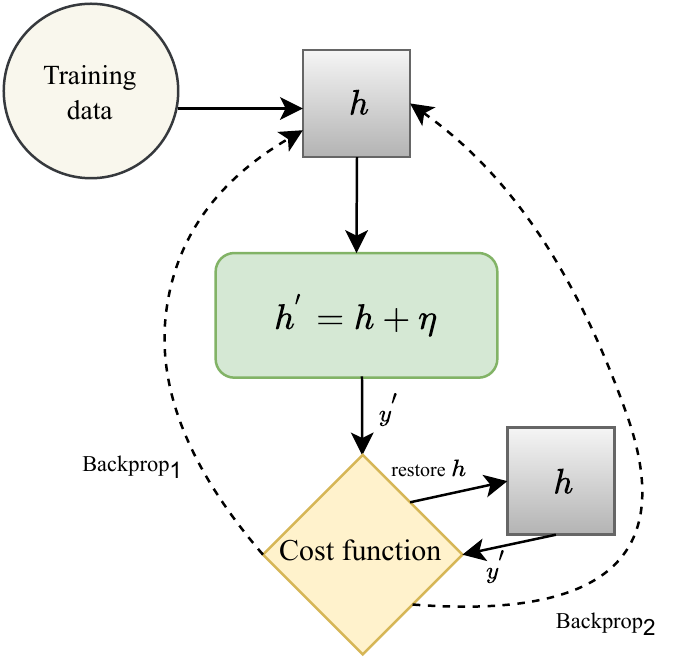}
\caption{Overall structure of FSGM. $y^{'}$ is the predicted label. $\eta$ is added noise at the point of interception $h$.}
\label{FGSMstructure}
\end{figure}

In our implementation (Figure \ref{FGSMstructure}), alongside the main component, there is an attacker that intercepts the model at a certain step during each training epoch, makes a copy of the pre-attack parameters in the intercepted layer, and injects noise into the model. Based on this information, an adversarial loss is computed and backpropagation is applied. 

After this step, a \texttt{restore} function is executed, returning the parameters of the intercepted layer back to its pre-attack values. A regular loss is then computed and backpropagation is applied for a second time. This added noise is computed based on equation \ref{eta}. If $h$ is the representation of a training instance at the time of interception by the attacker, the perturbation is calculated by $h^{'} = h+ \eta$.

\subsection{Dataset}
\label{sec:dataset}

For the experiments in this study we use a hospital dataset which we refer to as OUH. OUH is a de-identified EHR dataset, covering unscheduled emergency presentations to emergency and acute medical services at Oxford University Hospitals NHS Foundation Trust (Oxford, UK). These hospitals consist of four teaching hospitals, which serve a population of $600,000$ and provide tertiary referral services to the surrounding region. At the time of model development, linked deidentified demographic and clinical data were obtained for the period of November 30, 2017 to March 6, 2021. For each presentation, data extracted included presentation blood tests, blood gas results, vital sign measurements, results of RT-PCR assays for SARS-CoV-2, and PCR for influenza and other respiratory viruses. Patients who opted out of EHR research, did not receive laboratory blood tests, or were younger than 18 years of age have been excluded from this dataset. 

For OUH, hospital presentations before December 1, 2019, and thus before the global outbreak, were included in the COVID-19-negative cohort. Patients presenting to hospital between December 1, 2019, and March 6, 2021, with PCR confirmed SARS-CoV-2 infection, were included in the COVID-19-positive cohort. This period includes both the first and second waves of the pandemic in England \footnote{\url{https://coronavirus.data.gov.uk/details/cases}} . Because of incomplete penetrance of testing during early stages of the pandemic and limited sensitivity of PCR swab tests, there is uncertainty in the viral status of patients presenting during the pandemic who were untested or tested negative. Therefore, these patients were excluded from the datasets. 

\begin{figure}[ht]
\centering
\includegraphics[width=9.5cm]{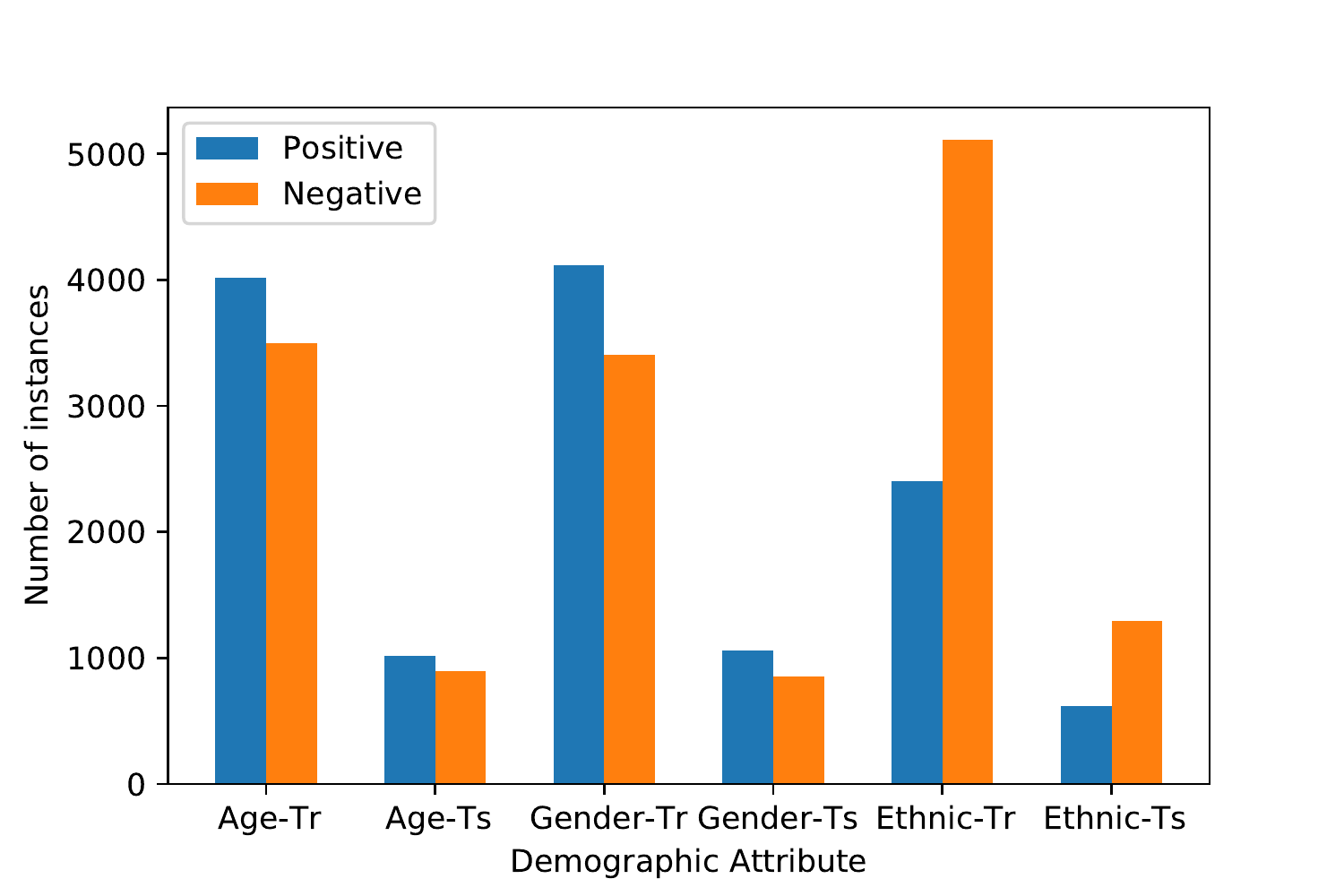}
\caption{Distribution of labels for each demographic attribute in TRAIN(-Tr) and TEST(-Ts) sets in OUH}
\label{class-dist}
\end{figure}

There are $3081$ instances of COVID-19-positive in the original dataset and $112121$ negative instances. For the experiments with OUH, we subsampled the majority class to reach a more balanced dataset with prevalence $0.5$ (i.e. $6162$ positive labels). Age, gender, and ethnicity information were binarised during preprocessing. For gender, the average age is $64$, which is taken as cut-off point for binarisation \footnote{There was not a big gap between median and mean, therefore we simply used average}. The ethnicity information, which were encoded using NHS ethnic categories, were divided into white and non-white. While quantising features in this way involves oversimplification and loss of detail, it keeps the values binary across all the protected attributes making comparisons easier in our experimental setup. Table \ref{class-dist} shows the distribution of demographic labels in the OUH dataset.  

In Section \ref{sec:external-valiation}, we will externally validate our models on three NHS Foundation Trust datasets \citep{soltan2022real}, namely Bedfordshire Hospitals NHS Foundation Trust (BH), University Hospitals Birmingham NHS Foundation Trust (UHB), and Portsmouth University Hospitals NHS Trust (PUH). We will use the entire test sets in their original label distribution within the pandemic time-frame to make sure the evaluation is fair and that it mirrors the highly imbalanced data used in hospitals. Table \ref{t:external-validation} shows the statistics for the Covid-19 Positive cases in the datasets. 

\begin{table}[ht]
\small
\centering
\smallskip
\caption{\label{t:external-validation} Label distributions for PCR (along with percentage of each label) for UHB, BH, and PUH datasets used for external validation of the models}
\begin{tabular}{c |cccc}
 & COVID+ & COVID- & total \\
UHB & $624$ ($1.48$\%)& $42095$ ($98.52$\%)& $42719$ \\
BH  & $209$ ($11.13$\%)& $1669$ ($88.87$\%)& $1878$ \\
PUH  & $2002$ ($5.2$\%)& $36579$ ($94.8$\%)& $38581$ \\
\hline
\end{tabular}
\end{table}

Evaluation at UHB trust considered all patients presenting to The Queen Elizabeth Hospital, Birmingham, between December 01, 2019 and October 29, 2020. The Queen Elizabeth Hospital is a large tertiary referral unit within the UHB group which provides healthcare services for a population of $2.2$ million across the West Midlands. Confirmatory COVID-19 testing was performed by laboratory SARS-CoV-2 RT-PCR assay.

Evaluation at BH considered all patients presenting to Bedford Hospital between January 1, 2021 and March 31, 2021. BH provides healthcare services for a population of around $620,000$ in Bedfordshire. Confirmatory COVID-19 testing was performed by point-of-care PCR based nucleic acid testing [SAMBA-II \& Panther Fusion System, Diagnostics in the Real World, UK, and Hologic, USA].

Evaluation at PUH considered all patients admitted to the Queen Alexandria Hospital, serving a population of $675,000$ and offering tertiary referral services to the surrounding region, between March 1, 2020 and February 28, 2021. Confirmatory COVID-19 testing was by laboratory SARS-CoV- 2 RT-PCR assay. 

\section{Experiments and Results}
\label{experiments-results}

We performed a series of experiments in order to test the proposed models and compare them against baselines. The baseline non-adversarial model that we use as the basic structure to start from, consists of $3$ fully connected dense layers with batch normalisation and dropout. We refer to this model as Base. During 10-fold cross-validation, the best hyperparameters were chosen using random search. We empirically found that heavy hyperparameter optimisation had at best mixed results and adding more layers to the model did not consistently boost performance. We chose a set of parameters that seemed to work well across all the models during cross-validation (Table \ref{tab:hyperparams}) \footnote{All the models with the exception of ADV\textsubscript{per} were trained for $15$ epochs for experiments on OUH. For external validation, this was set to $30$ epochs. ADV\textsubscript{per} seemed to require more training epochs in all the experiments, therefore we trained it with $30$ epochs for both OUH and external validation.}. We also kept the Base model simple with only a few layers so we could have direct and straightforward comparisons with the adversarially trained models. The demographic-based adversarial model is referred to as ADV and its main component is the same as Base. Since after training, only the Base part will be tested (i.e. discriminators will detach), the ADV model ends up having the exact same number of parameters as Base. The perturbation-based adversarial model, which also has the same number of parameters as Base, is referred to as Adv\textsubscript{per}. All the reported results on the test set are the median of three consecutive runs. 

\begin{table}[ht]
\setlength\tabcolsep{1.5pt}
\caption{Hyperparameter values used for all the experiments}
\label{tab:hyperparams}
\centering
\begin{tabular}{|c|c|c|c|c|c|c|c|}
\hline
learning rate   & $\lambda$   & batch size    & hidden dimension (Base)   & hidden dimension (disc)   & dropout   & epochs    \\ \hline
$0.0008$   &   $2$     & $16$          & $150$                     & $300$                     & $0.5$     & $15$/$30$             \\ \hline

\end{tabular}
\end{table}

In what follows we explain the feature sets used, the train and test procedure and finally report the main task and attacker results under different scenarios. 

\subsection{Feature sets}
Two sets of clinical variables were investigated (Table~\ref{feats-tab}): presentation blood tests from the first blood draw on arrival to hospital and vital signs. Only blood test markers that are commonly taken within existing care pathways and are usually available within 1 hour in middle and high-income countries were considered here. 

\begin{table}[ht]
\caption{Clinical parameters included in each feature set}
\label{feats-tab}
\begin{tabular}{|p{4cm}|p{9cm}|}
\hline
{Feature Type} & {Features included} \\ \hline
\hline
Presentation blood tests &	Haemoglobin, haematocrit, mean cell volume, white cell count, neutrophil count, lymphocyte count, monocyte count, eosinophil count, basophil count, platelets, prothrombin time, INR, APTT, sodium, potassium, creatinine, urea, eGFR, C Reactive Protein (CRP), albumin, alkaline phosphatase, ALT, bilirubin \\ \hline \hline
Presentation vital signs & 	Heart rate, respiratory rate, oxygen saturation, systolic blood pressure, diastolic blood pressure, temperature, oxygen flow rate \\ \hline 
\end{tabular}
\end{table}

\subsection{Training and testing}
The models are trained and tested in a binary classification task in which the labels are confirmed PCR test results. As the first step, the model is evaluated on the TRAIN set in a stratified 10-fold cross-validation scenario during which a threshold is set on the ROC curve to meet the minimum recall constraint \footnote{The idea behind calibration of recall is to make sure the false negatives do not exceed beyond a certain point. In a hospital setting and in a pandemic, it is too costly to send patients home with a false negative result or transfer them to wards and potentially expose other inpatients to infection. Therefore, high sensitivity is needed to give physicians confidence that negative results are truly negative.}. Consequently, the model is trained on the TRAIN set and tested on the holdout TEST data and results are computed using the previously set threshold.

\begin{figure}[ht]
\centering
\includegraphics[width=9.5cm]{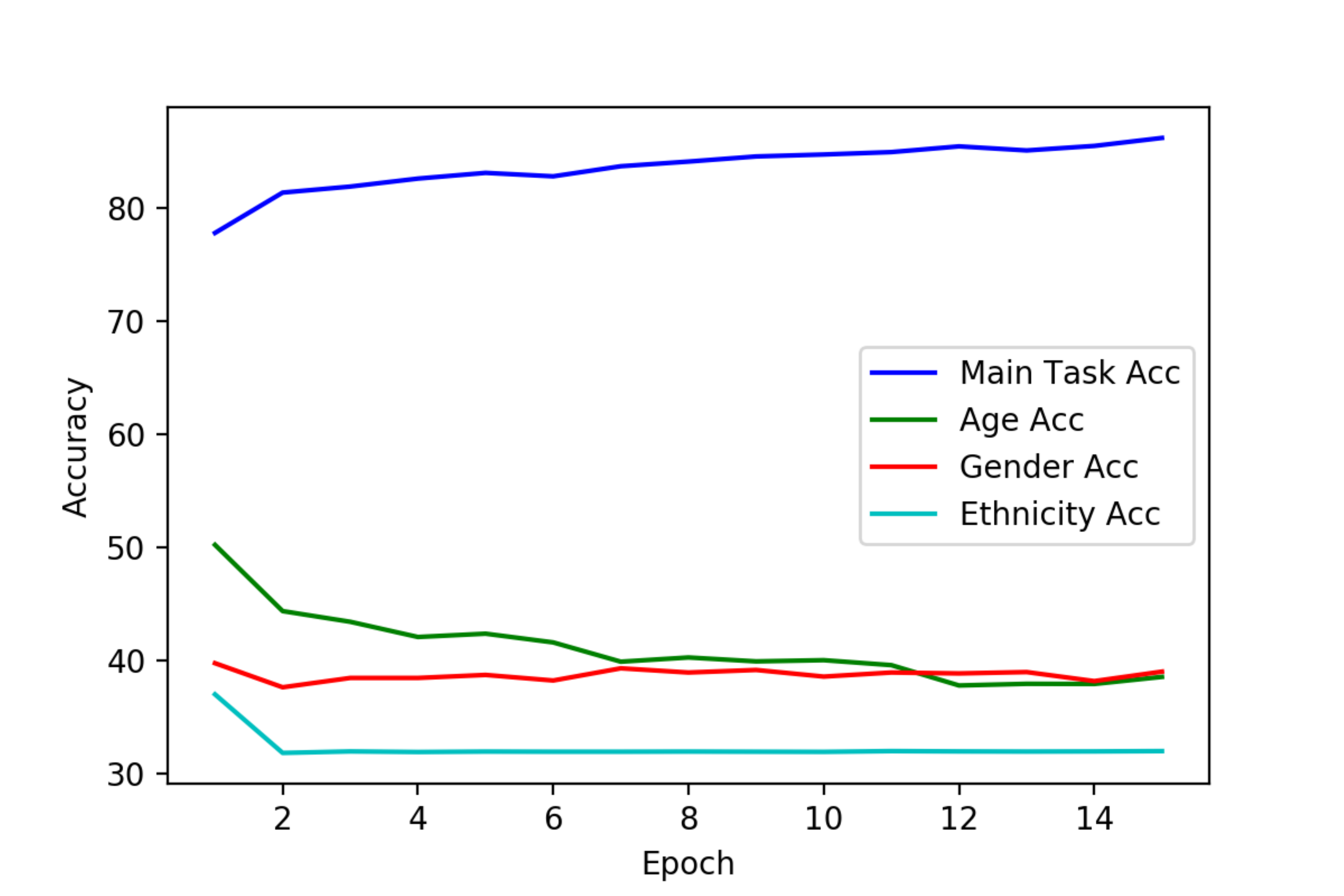}
\caption{Accuracy scores for the main and each of the three discriminators for each epoch}
\label{acc-epoch}
\end{figure}

During training of the ADV model, the expectation is that the accuracy of the main classifier increase over subsequent epochs, and since the learning setup is such that discriminators are constantly misled, performance is intended to be kept below or around $50\%$ accuracy. To test this assumption, we plotted the changes in the trajectory of accuracy for the main and three auxiliary tasks in the first $15$ epochs. This is when the ADV model is being trained on TRAIN set and before it is tested on holdout TEST. As can be seen in Figure \ref{acc-epoch}, accuracy for the main task keeps growing steadily while discriminator accuracy drops below $50\%$ and plateaus afterwards.       

In Table \ref{tab:main-results-cv} we report the results on the main task of predicting PCR results for all the models. The results demonstrate the models perform well at the main task, namely, predicting the outcome of the PCR test. 

\begin{table}[ht]
\caption{Results for the $4$ different models for the OUH dataset in a 10-fold cross-validation setting with the recall value set as $0.8\pm0.07$}
\label{tab:main-results-cv}
\setlength\tabcolsep{2pt}
\begin{tabular}{|c||c|c|c|c|c|c|c|c|c|}
\hline
Model        & Recall & Precision & F1-Score & Accuracy & Specificity & PPV    & NPV    & AUC    & Threshold               \\ \hline
Base         & 0.7335 & 0.7356    & 0.7341   & 0.8219   & 0.8670      & 0.7356 & 0.8667 & 0.8623 & 0.1551                  \\ \hline
ADV\textsubscript{per} & 0.7331 & 0.7355    & 0.7338   & 0.8216   & 0.8668      & 0.7355 & 0.8665 & 0.8571 & 0.0464        \\ \hline
ADV          & 0.7355 & 0.7308    & 0.7325   & 0.8199   & 0.8629      & 0.7308 & 0.8670 & 0.8553 & 0.1669                  \\ \hline
\end{tabular}
\end{table}

\subsection{Attacking trained networks to predict protected attributes}
\label{attacking-models}

In order to asses how much privacy each model can provide against an adversarial attack, we perform a series of experiments in which $3$ different non-adversarial Base models are trained on the training data, with each corresponding to the prediction of a different demographic attribute. In other words, instead of predicting the PCR test result, a protected attribute is provided as the label to train and test on. We perform the experiments under the same conditions as the main task. The attacker is first trained in a 10-fold cross-validation scenario and a threshold is set based on the ROC curve with the minimum recall constraint of $0.8\pm0.07$. 

Subsequently, the attackers are trained on TRAIN set and tested on the TEST portion of the dataset and predict the same values given the obtained threshold set during 10-fold CV. These results are important to the final interpretations of the model privacy because they determine the upper bound for the most amount of leak the proposed models can have. In Table \ref{attacker-results}, we report the results for trained attackers on the TEST portion of the dataset given each protected attribute that was predicted. 

The lower bound is the the majority class baselines in which the attacker simply relies on some prior information about the distribution of the protected attributes to predict these features and does not make use of the obtained hidden representations. For instance, if a dataset is obtained in Scotland, relying on the known fact that the predominant ethnic category is British White, the attacker would simply assign the same label to all of the instances. Statistics about majority classes for each attribute is given in Table \ref{majority-baselines} in both TRAIN and TEST sets. As can be seen, ethnicity is the most unbalanced category in comparison with gender and age in which class labels are more equally distributed.

\begin{table}[ht]
\caption{Attacker results on the TEST set when trained and tested on features directly. This serves as the upper bound for information leakage}
\label{attacker-results}
\setlength\tabcolsep{2pt}
\begin{tabular}{|c||c|c|c|c|c|c|c|c|}
\hline
Predicted Attribute        & Recall & Precision & F1-Score & Accuracy & Specificity & PPV     & NPV     & AUC    		\\ \hline
Age          			   & 0.7193 & 0.7470    & 0.7329   & 0.7936   & 0.8782      & 0.74704 & 0.8622  & 0.8884   	    \\ \hline
Gender       			   & 0.7346 & 0.78897   & 0.7608   & 0.8092   & 0.9017      & 0.7889  & 0.8717  & 0.9104    	\\ \hline
Ethnicity				   & 0.6688 & 0.4506    & 0.5384   & 0.6170   & 0.5922      & 0.4506  & 0.7815  & 0.6737 		\\ \hline

\end{tabular}
\end{table}

\begin{table}[ht]
\setlength\tabcolsep{2pt}
\caption{Percentage of majority class labels to the whole data for each demographic attribute}
\label{majority-baselines}
\centering
\begin{tabular}{|c||c|c|}
\hline
Protected attribute   & TRAIN & TEST  \\ \hline
Age                   & 0.53 & 0.53   \\ \hline
Gender                & 0.54 & 0.55   \\ \hline
Ethnicity             & 0.68 & 0.67   \\ \hline
\end{tabular}
\end{table}

As the next step, we trained our baseline and proposed adversarial models on the TRAIN set and saved the weights of the neural networks. We then loaded our trained attackers and tested the attackers, not on the feature directly this time, but on the output of the encoder of the baseline and adversarially trained models. The idea is that, if an adversarially trained model is indeed protecting demographic attributes, it should make it harder for an attacker to predict those values from its encoded representations in comparison with a baseline model that is not specifically designed for preservation of privacy. Results shown in Table \ref{attacker-results-base-encoder} already show a degree of privacy provided by the non-adversarial encoder, as they indicate a noticeable decrease in performance compared to Table \ref{attacker-results}. The most marked decrease is visible in prediction of gender, in which performance drops from AUC of $0.9104$ to $0.6926$. In the case of age, however, the attacker seems more robust.

\begin{table}[ht]
\caption{Attacker results on the TEST set when trained and tested on the output generated by the encoder of the non\-adversarial Base model}
\label{attacker-results-base-encoder}
\setlength\tabcolsep{2pt}
\begin{tabular}{|c||c|c|c|c|c|c|c|c|}
\hline
Predicted Attribute        & Recall & Precision & F1-Score & Accuracy & Specificity & PPV     & NPV     & AUC    		\\ \hline
Age          			   & 0.7831 & 0.5855 	& 0.6701   & 0.7549   & 0.7229      &0.5855   &0.8695   & 0.8131  	    \\ \hline
Gender       			   & 0.7969 & 0.4336    & 0.5616   & 0.6551   & 0.4795      &0.4336   &0.8252   & 0.6926   		\\ \hline
Ethnicity				   & 0.7835 & 0.3776    & 0.5096   & 0.4932   & 0.3544      & 0.3776  & 0.7660  & 0.6265		\\ \hline

\end{tabular}
\end{table}

Since we want to keep the attackers blind to the encoding strategy used by the adversarially trained model, in order to test the attackers on the ADV and ADV\textsubscript{per} models, we have to use the same threshold set during 10-fold CV on the encoded representation of the Base model. Therefore, we load the attacker which is trained on the non-adversarial encoder on the TRAIN set and test it on the ADV/ADV\textsubscript{per} model's encoder to predict the three attributes.  

\begin{table}[ht!]
\caption{Attacker results on the TEST set when trained on the encoder of the Base model and tested on the encoder of the ADV model}
\label{attacker-results-adv-encoder}
\setlength\tabcolsep{2pt}
\begin{tabular}{|c||c|c|c|c|c|c|c|c|}
\hline
Predicted Attribute        & Recall & Precision & F1-Score & Accuracy & Specificity & PPV     & NPV     & AUC    		\\ \hline
Age          			   & 0.8213 & 0.3439 	& 0.4849   & 0.5386   & 0.2167      & 0.3439  & 0.7082  & 0.5744 	    \\ \hline
Gender       			   & 0.6676 & 0.3117    & 0.4250   & 0.4869   & 0.2631      & 0.3117  & 0.6129  & 0.4572  		\\ \hline
Ethnicity				   & 0.4394 & 0.3493    & 0.3892   & 0.5417   & 0.5907      & 0.3493  & 0.6782  & 0.5112		\\ \hline

\end{tabular}
\end{table}

The results in Tables \ref{attacker-results-adv-encoder} and \ref{attacker-results-perturb-encoder} confirm the assumption that an adversarial learning procedure, either with separate discriminator networks for each protected attribute or using perturbation-based regularisation, provides a greater level of privacy against attacks by an intruder that intends to recover this information using a representation obtained from the model. 

\begin{table}[ht!]
\caption{Attacker results on the TEST set when trained on the encoder of the Base model and tested on the encoder of the ADV\textsubscript{per} model}
\label{attacker-results-perturb-encoder}
\setlength\tabcolsep{2pt}
\begin{tabular}{|c||c|c|c|c|c|c|c|c|}
\hline
Predicted Attribute        & Recall & Precision & F1-Score & Accuracy & Specificity & PPV     & NPV     & AUC    		\\ \hline
Age          			   & 0.4946 & 0.3164    & 0.3859   & 0.4811   & 0.4659      & 0.3164  & 0.64835 & 0.4723 	    \\ \hline
Gender       			   & 0.6421 & 0.3269    & 0.4333   & 0.5067   & 0.3391      & 0.3269  & 0.6546  & 0.5189  		\\ \hline
Ethnicity				   & 0.4151 & 0.3395    & 0.37355  & 0.5376   & 0.5961      & 0.3395  & 0.6709  & 0.4870		\\ \hline

\end{tabular}
\end{table}

\subsubsection{Demographic cross-testing to asses generalisability}
\label{sec:demographic-crosstest}

The application of an adversarial learning procedure to protect selected attributes involves a training setup with competing losses which is intended to weaken undesirable implicit associations contained in the hidden representations of the network. This is expected to result in a certain amount of performance drop compared to the non-adversarial baseline. As long as this drop is not massive, the performance-privacy trade-off is justified. However, a more general concern is whether a model like ADV, with its $3$ different discriminators and the direct and specific manipulation of its hidden representations would generalise poorly when tested on certain demographic sub-populations of the dataset. Since ADV\textsubscript{per} applies its regularisation without specifically targeting any protected attributes, it is less likely to suffer from this issue. 

In order to investigate whether protecting demographic attributes damages generalisability of the ADV, we performed a series of experiments with the aim to train and test our Base and ADV models only on one demographic group and tested it on the other. We compare the adversarial model with the baseline to make sure that generalisability of the ADV model is not hurt. Since we have $3$ different binary attributes, there are $6$ possible ways to cross-test the models. We denote these subgroups with f (female), m (male), w (white), n (non-white), o (old), and y (young) \footnote{Old and young here are simply labels to distinguish the two age sub-groups and do not necessarily reflect notions of young and old in society}. To restructure the dataset for these experiments, in each case we combine all the data and filter TRAIN and TEST based on the targeted demographic. For example `m2f' would mean that our TRAIN set only contains females and the TEST set only males. The results in Table \ref{demographic-crosstest} clearly indicate that adversarial learning has not damaged generalisability in any of scenarios in which the Base and ADV models were tested.  

\begin{table}[ht]
\caption{Results of demographic cross-tests to assess the effects of adversarial training on generalisability across different subgroups of the dataset. }
\label{tab:generalisability}
\label{demographic-crosstest}
\setlength\tabcolsep{3pt}
\begin{tabular}{|c|c||c|c|c|c|c|c|c|c|}
\hline
Cross-test   & Model       & Recall & Precision & F1-Score & Accuracy & Specificity & PPV     & NPV     & AUC   		\\ \hline
f2m		 	 & Base        & 0.6876 & 0.7401    & 0.7100   & 0.8119   & 0.8768      & 0.7401  & 0.8501  & 0.8435		\\ \hline
f2m		 	 & ADV         & 0.7130 & 0.7310    & 0.7191   & 0.8137   & 0.8663      & 0.7310  & 0.8591  & 0.8452		\\ \hline \hline
m2f		 	 & Base		   & 0.6720 & 0.7520    & 0.7085   & 0.8145   & 0.8889      & 0.7520  & 0.8449  & 0.8403		\\ \hline
m2f		 	 & ADV		   & 0.6947 & 0.7405    & 0.7142   & 0.8137   & 0.8759      & 0.7405  & 0.8527  & 0.8389		\\ \hline \hline
n2w			 & Base		   & 0.6771 & 0.7428    & 0.7057   & 0.8115   & 0.8813      & 0.7428  & 0.8466  & 0.8406		\\ \hline
n2w			 & ADV 		   & 0.6971 & 0.7281    & 0.7105   & 0.8101   & 0.8687      & 0.7281  & 0.8524  & 0.8397		\\ \hline \hline
w2n			 & Base        & 0.6802 & 0.7442    & 0.7077   & 0.8126   & 0.8815      & 0.7442  & 0.8479  & 0.8424		\\ \hline
w2n 		 & ADV         & 0.7037 & 0.7302    & 0.7149   & 0.8122   & 0.8686      & 0.7302  & 0.8552  & 0.8428		\\ \hline \hline
o2y          & Base        & 0.6873 & 0.7449    & 0.7123   & 0.8140   & 0.8800      & 0.7449  & 0.8502  & 0.8449		\\ \hline
o2y          & ADV         & 0.7019 & 0.7344    & 0.7153   & 0.8131   & 0.8709      & 0.7344  & 0.8550  & 0.8435		\\ \hline \hline
y2o          & Base        & 0.6716 & 0.7425    & 0.7021   & 0.8098   & 0.8817      & 0.7425  & 0.8445  & 0.8379		\\ \hline
y2o          & ADV         & 0.6922 & 0.7238    & 0.7055   & 0.8066   & 0.8660      & 0.7238  & 0.8501  & 0.8364		\\ \hline		   

\end{tabular}
\end{table}

\subsubsection{External Validation of the Models}
\label{sec:external-valiation}

In order to validate the models on external data, we trained Base, ADV, and ADV\textsubscript{per} on the OUH dataset (as described in Section \ref{sec:dataset}) and tested it on the entirety of the UHB, BH, and PUH datasets. We performed the same procedure as the previous experiments: First we ran a 10-fold CV on the OUH dataset and set a threshold and then tested the models on the external test data with the previously obtained threshold. The hyperparameters were kept the same for these experiments with the exception of ADV\textsubscript{per} which seemed to converge better after $30$ epochs during 10-fold CV. Tables \ref{tab:external-validation}, \ref{tab:external-validation-bedford}, and \ref{tab:external-validation-portsmouth} show the results of this experiment on the UHB, BH, and PUH test sets, respectively. 

\begin{table}[ht]
\caption{Results for the models when trained on OUH and tested on the UHB dataset }
\label{tab:external-validation}
\setlength\tabcolsep{5.5pt}
\begin{tabular}{|c||c|c|c|c|c|c|c|c|}
\hline
Model        & Recall & Precision & F1-Score & Accuracy & Specificity & PPV    & NPV    & AUC                   \\ \hline
Base         & 0.7371 & 0.7261	  & 0.7316	 & 0.8602	& 0.8609	  & 0.7261 & 0.8675	& 0.8643	            \\ \hline
ADV\textsubscript{per} & 0.7155	 & 0.7286	 & 0.7218	& 0.8657	 & 0.86669 & 0.7286	& 0.8591	& 0.8531         \\ \hline
ADV          & 0.7275 & 0.7236	 & 0.7256	 & 0.8602	& 0.8611	 & 0.7236  & 0.8634	& 0.8586                   \\ \hline
\end{tabular}
\end{table}

\begin{table}[ht]
\caption{Results for the models when trained on OUH and tested on the BH dataset }
\label{tab:external-validation-bedford}
\setlength\tabcolsep{5.5pt}
\begin{tabular}{|c||c|c|c|c|c|c|c|c|}
\hline
Model        & Recall & Precision & F1-Score & Accuracy & Specificity & PPV    & NPV    & AUC                   \\ \hline
Base         & 0.6556 & 0.7760	  & 0.7045	 & 0.8795	& 0.9002	 & 0.7760 & 0.8414	& 0.8608				\\ \hline
ADV\textsubscript{per} & 0.6301	 & 0.7562	 & 0.6767	& 0.8746	 & 0.8933 & 0.7562	& 0.8320 &	0.8115      \\ \hline
ADV          & 0.6923 & 0.7473	  & 0.7163	 & 0.8690	& 0.8806	 & 0.7473 & 0.8521	& 0.8506
                  \\ \hline
\end{tabular}
\end{table}

\begin{table}[ht]
\caption{Results for the models when trained on OUH and tested on the PUH dataset }
\label{tab:external-validation-portsmouth}
\setlength\tabcolsep{5.5pt}
\begin{tabular}{|c||c|c|c|c|c|c|c|c|}
\hline
Model        & Recall & Precision & F1-Score & Accuracy & Specificity & PPV    & NPV    & AUC                   \\ \hline
Base         & 0.6988 & 0.7441    & 0.7168   & 0.8638   & 0.8762      & 0.7441 & 0.8545 & 0.8567				\\ \hline
ADV\textsubscript{per} & 0.6401   & 0.7575   & 0.6858   & 0.8768      & 0.8937 & 0.7575 & 0.8351 & 0.8173       \\ \hline
ADV          & 0.6973  & 0.7450   & 0.7184   & 0.8680   & 0.8788      & 0.7450 & 0.8537 & 0.8527
                  \\ \hline
\end{tabular}
\end{table}

\section{Discussion and Conclusion}
\label{sec:discusson}

In this work, we introduced and tested two adversarially trained models for the task of predicting COVID-19 PCR test results based on routinely collected blood tests and vital signs. The data was processed in the form of tabular data. 

In our experiments, we addressed the issue of leakage of potentially sensitive attributes that are implicitly contained in the dataset, and demonstrated how an attacker network can successfully retrieve this information under different circumstances. Information like age seem to be easily inferred with high accuracy from the features or from the hidden representation of the Base model. In this case, ADV and ADV\textsubscript{per} models significantly reduced this vulnerability, which highlights the protective power of these adversarial methods in hiding such implicit information against invasive models that are specifically trained to infer this knowledge. 

The same pattern was seen in the case of the other two demographic attributes, namely, gender and ethnicity. For ethnicity, the representation was less informative to the attacker network for the following two reasons: 

\begin{enumerate}
  \item A certain percentage of the patients had preferred not to state their ethnicity. Since we wanted to keep all the tasks binary, we treated this category as non-white which is clearly sub-optimal. This further complicates ethnicity prediction for the attacker.
  \item There are limitations in the accuracy of documenting ethnicity by hospital staff during data collection, which may increase the amount of noise in the data.
\end{enumerate}

However, even though the overall results are lower for the case of ethnicity, the ADV model still shows better privacy compared to the baseline. In such cases, the adversary is likely to rely on prior knowledge of the dataset or general information about the prevalence of ethnicity groups in the data, rather than the output of the encoder. 

Our adversarial setup came with only a minimal performance cost (Table \ref{tab:main-results-cv}) and proved robust both in the generalisability tests (Table \ref{tab:generalisability}) and in external validation on highly imbalanced datasets (Section \ref{sec:external-valiation}). More experiments (both at the level of data and model) are needed to ascertain whether the same general patterns can be seen under different conditions. Nonetheless, these methods are not tied to the specifics of the Base model and can be applied to any neural architecture. Furthermore, in the case of the ADV model, the protected attributes need not be demographic and theoretically any categorical feature of interest (or any feature that can be meaningfully quantised) can be used during training. Future work can also include experimenting with continuous features, in which the attacker would have to guess the features in a regression task.

To conclude, in this paper we introduced two effective methods to protect sensitive attributes in a tabular dataset related to the task of predicting COVID-19 PCR test result based on routinely collected clinical data. We demonstrated the effectiveness of adversarial training by assessing the proposed models against a comparable baseline both in the context of the main task where it showed performance scores that were by and large at the same level with the baselines and also in the context of privacy preservation where a trained attacker was employed to retrieve sensitive information by intercepting the content of the models' encoder. In the second scenario, the adversarially trained models consistently showed superior performance compared to the non-adversarial baseline. 

\section*{Acknowledgement}

This work uses data provided by patients and collected by the UK’s National Health Service as part of their care and support. We thank all the people of Oxfordshire who contribute to the Infections in Oxfordshire Research Database.

Research Database Team: L Butcher, H Boseley, C Crichton, DW Crook, D Eyre, O Freeman, J Gearing (community), R Harrington, K Jeffery, M Landray, A Pal, TEA Peto, TP Quan, J Robinson (community), J Sellors, B Shine, AS Walker, D Waller. Patient and Public Panel: G Blower, C Mancey, P McLoughlin, B Nichols.

NHS Health Research Authority (HRA) approval was granted for the use of routine clinical and microbiology data from Electronic Health Records for development and validation of artificial intelligence models to detect Covid-19 (CURIAL; IRAS ID 281832).

We additionally express our gratitude to Jingyi Wang \& Dr Jolene Atia at University Hospitals Birmingham NHS Foundation trust, Phillip Dickson at Bedfordshire Hospitals, and Paul Meredith at Portsmouth Hospitals University NHS Trust for assistance with data extraction.

\section*{Funding}

This study was supported by the Wellcome Trust/University of Oxford Medical \& Life Sciences Translational Fund (Award: 0009350) and the Oxford National Institute of Research (NIHR) Biomedical Research Campus (BRC). The funders of the study had no role in study design, data collection, data analysis, data interpretation, or writing of the manuscript. AS is an NIHR Academic Clinical Fellow. The views expressed are those of the authors and not necessarily those of the NHS, NIHR, or the Wellcome Trust.

\bibliography{iclr2021_conference}
\bibliographystyle{iclr2021_conference}


\end{document}